# Deep Learning-based Method for Expressing Knowledge Boundary of Black-Box LLM

Haotian Sheng*, Heyong Wang, Ming Hong, Hongman He and Junqiu Liu[1]

*Abstract*—**Large Language Models (LLMs) have achieved remarkable success; however, the emergence of content generation distortion (hallucination) limits their practical applications. The core cause of hallucination lies in LLMs' lack of awareness regarding their stored internal knowledge, preventing them from expressing their knowledge state on questions beyond their internal knowledge boundaries, as humans do. However, existing research on knowledge boundary expression primarily focuses on white-box LLMs, leaving methods suitable for black-box LLMs— which offer only API access without revealing internal parameters—largely unexplored. Against this backdrop, this paper proposes LSCL (LLM-Supervised Confidence Learning), a deep learning-based method for expressing the knowledge boundaries of black-box LLMs. Based on the knowledge distillation framework, this method designs a deep learning model. Taking the input question, output answer, and token probability from a black-box LLM as inputs, it constructs a mapping between the inputs and the model's internal knowledge state, enabling the quantification and expression of the black-box LLM's knowledge boundaries. Experiments conducted on diverse public datasets and with multiple prominent black-box LLMs demonstrate that LSCL effectively assists black-box LLMs in accurately expressing their knowledge boundaries. It significantly outperforms existing baseline models on metrics such as accuracy and recall rate. Furthermore, considering scenarios where some black-box LLMs do not support access to token probability, an adaptive alternative method is proposed. The performance of this alternative approach is close to that of LSCL and surpasses baseline models.**

## I. INTRODUCTION

L ARGE Language Models (LLMs), such as GPT-4 [1] and DeepSeek-V3 [2], store extensive knowledge within their parameters, demonstrating remarkable performance across a wide range of language tasks. For example, LLMs excel at passing various examinations requiring specialized medical knowledge [3]and financial expertise [4]. However, the frequent occurrence of hallucinations in LLMs—where model outputs deviate from objective facts—has led to a marked reluctance among people to seek professional advice from them [5], [6]. The core cause of hallucination lies in LLMs' lack of awareness regarding their stored internal knowledge. This prevents them from expressing their knowledge state on questions beyond their internal knowledge boundaries, as humans do[7]—essentially, a deficiency in accurately expressing knowledge boundaries. Consequently, enhancing LLMs' awareness of their knowledge boundaries has become crucial for mitigating hallucinations and ensuring model reliability. Existing research on knowledge boundaries primarily focuses on fine-tunable white-box LLMs, with less attention paid to black-box LLMs that offer only API access without revealing internal parameters. Given that black-box LLMs typically possess larger parameter scales and broader knowledge bases, the effective identification of their knowledge boundaries is vital for expanding the scope of reliable LLM applications. Therefore, this paper focuses on the problem of knowledge boundary expression for black-box LLMs.

Expressing the knowledge boundaries of black-box LLMs faces two core challenges. The first challenge is how to achieve accurate knowledge boundary expression for black-box LLMs without accessing their internal parameters. Existing methods for expressing knowledge boundaries in white-box LLMs [8], [9], [10] leverage fine-tuning or internal signals to enable LLMs to express their knowledge boundaries, typically by responding with "Know"/"Unknow" or refusing to answer. Such approaches are inapplicable to black-box LLMs. The second core challenge lies in the fact that the widely adopted "Know/Unknow" binary classification paradigm inadequately captures the nuanced knowledge state of LLMs. This paradigm aims to align the model's knowledge state with the correctness of its answers. However, this classification scheme exhibits significant shortcomings for answers involving uncertainty. A key illustration of this is when an LLM outputs a correct answer during knowledge boundary evaluation using token probability, yet these token probabilities are low. Such low-probability correct answers reflect the model's uncertainty in responding to the query, indicating that inconsistent results would be produced across multiple samplings. Consequently, classifying such an answer as "Know" overlooks the model's inherent uncertainty, while classifying it as "Unknow" contradicts the alignment with answer correctness. Given the prevalence of such misclassification of low-probability correct answers, this leads to erroneous assessments of LLM knowledge boundary expression and undermines the practical utility of evaluation results for deployment.

To address the core challenges in expressing knowledge boundaries for black-box LLMs outlined above, this paper proposes LSCL (LLM-Supervised Confidence Learning), a deep learning-based method for LLM knowledge boundary expression. To tackle the first challenge, LSCL employs a deep learning framework inspired by knowledge distillation. This framework is designed to capture the LLM's knowledge state regarding the internal knowledge relevant to a given query, through the mapping between the model's inputs (questions) and outputs (answers and token probability). To address the second challenge, LSCL categorizes the LLM's knowledge state into three distinct classes: "Know", "Sciolism" and, "Unknow". The newly introduced "Sciolism" class is specifically designated for answers expressed by uncertainty.

[1] Haotian Sheng*, Heyong Wang, Ming Hong, Hongman He and Junqiu Liu are with the Department of Electronic Business, South China University of Technology, Guangzhou, Guangdong, China (e-mail: ebshenghaotian@mail.scut.edu.cn; wanghey@scut.edu.cn; ming@scut.edu.cn; 202410192491@mail.scut.edu.cn; ebliujunqiu@mail.scut.edu.cn ).

The core contributions of this paper are summarized as follows:

1) We propose LSCL, a knowledge boundary expression method suitable for black-box LLMs. Crucially, LSCL introduces an innovative deep learning-based confidence learning module. This module is dedicated to learning the mapping relationship between the internal knowledge stored within the parameters of a black-box LLM and the knowledge required to correctly answer specific tasks, thereby achieving accurate prediction of confidence scores. Since confidence calculation relies on the answers generated by the LLM in response to input questions, it inherently involves the two core elements: "question" and "answer." To address this dependency, we design and incorporate a question-answer alignment component. This component enables the confidence learning module to precisely capture both the local semantic associations and the global semantic consistency between the question and its corresponding answer. Furthermore, LSCL exhibits a significant advantage in low computational overhead compared to traditional methods that require fine-tuning or updating LLM parameters. Notably, the training and deployment of its confidence learning module were successfully conducted on a consumer-grade GPU (NVIDIA RTX 4060-Ti- 16GB). This demonstrates the engineering practicality of the proposed method.

2) We propose a novel confidence quantification metric—Correctness-Adjusted Token Probability (CATP). This metric enables precise quantification of an LLM's grasp of the knowledge required to answer specific tasks. Crucially, it transcends the conventional binary paradigm prevalent in prior research, which simplistically categorized knowledge states as merely "Know" or "Unknow." Instead, our metric effectively distinguishes and expresses three distinct knowledge states: "Know", "Unknow," and the intermediate state of "Sciolism." This tripartite distinction provides fundamental support for refined knowledge boundary expression.

3) LSCL exhibits superiority in comparative experiments. Comparative experiments were conducted on various mainstream LLMs, demonstrating that the proposed LSCL method significantly outperforms existing baseline models. The experimental results fully validate that LSCL can effectively learn and model the internal knowledge representation of black-box LLMs, achieving precise and clear expression of their knowledge boundaries. This provides key technical support for the reliable deployment of black-box LLMs.

## II. Literature Review

Given that confidence calibration and knowledge boundary expression are closely aligned with the core objectives of this study, we focus our review and discussion on these two categories of methods in the following sections.

### A. Confidence Calibration Methods for Black-box LLMs

In our work, the calibrated confidence of black-box LLMs serves as a key variable for assessing knowledge mastery states. In this sense, our research is related to the literature on confidence calibration methods for black-box LLMs. Confidence calibration methods can be divided into two main categories: confidence estimation methods and confidence expression methods, specifically as follows: (1) Confidence estimation methods: These methods often employ prediction probability as the core metric for LLM confidence, with token probability being a fundamental estimation approach—it models the probability of tokens generated under a given context based on the parametric knowledge of LLMs. Generally, high token probability indicates that the model has high confidence in the generated answer, implying sufficient mastery of the knowledge required to address the question; conversely, low token probability suggests insufficient model confidence, reflecting uncertainty in internal knowledge. However, token probability is limited by its reliance on a single decoding strategy. This section elucidates the relationship between the current study and confidence calibration research. To address this, researchers have proposed confidence estimation methods compatible with multiple decoding strategies. Examples include: estimating confidence by statistically measuring the consistency frequency of sampled outputs from multiple runs with identical inputs [11], [12] or directly leveraging LLMs to probabilistically assess the correctness of their own predictions [13]. Additional calibration enhancement techniques include prompt ensemble [14], hybrid methods [15], fidelity evaluation [16], and model ensemble [17], [18]. (2) Confidence expression methods: These approaches focus on guiding LLMs to directly incorporate confidence indicators as tokens within their predictions, enabling simultaneous output of confidence levels and answers [19], [20], [21]. For instance, Tian et al. [20] introduced the "Guiding by Prompt" method, which employs specially designed prompts to direct LLMs to output tokens representing their confidence levels while generating answers, thereby directly reflecting the model's knowledge mastery state.

### B. Knowledge Boundary Expression in White-Box Large Language Models

Unlike confidence calibration which quantifies uncertainty, knowledge boundary expression focuses on assessing the knowledge mastery state regarding question-answer capabilities. Current research in LLM knowledge boundary expression generally posits that LLMs cannot autonomously articulate their ignorance [7], precluding the direct use of prompt-guided methods. Consequently, such research necessitates fine-tuning or signal probing techniques, thus applying only to white-box LLMs. For example, R-tuning [10] leverages annotated data to evaluate answer correctness and employs supervised fine-tuning to achieve knowledge boundary expression. COKE [8] demarcates knowledge boundaries by probing internal confidence levels across question sets, then induces explicit boundary articulation based on probing results. Kang et al. [9] explore reinforcement learning fine-tuning to guide LLMs in proactively responding "I don't know" to unfamiliar questions. Furthermore, it should be noted that existing studies uniformly simplify LLM knowledge boundary expression into a binary classification framework, where the "Know" state indicates complete and accurate knowledge for answering, while the "Unknow" state signifies incomplete possession of required knowledge.

### C. Critical Literature Review

Compared to existing confidence calibration research, the innovations of this study primarily manifest in two key aspects: On one hand, a novel confidence estimation method is proposed as the prediction target for the introduced model. This method is guided by the core objective of confidence calibration—aligning confidence with prediction accuracy—and integrates two essential elements: the correctness of LLM responses and token probability, thereby achieving precise quantification of the knowledge mastery degree required for task-specific answering. Existing research has demonstrated the suboptimal calibration performance of traditional token probability [20], [22], with core defects evident in two aspects: First, token probability exhibits orthogonality to answer correctness, as its value depends solely on the model's internal knowledge generation; even when the response is erroneous, it may still yield high probability values. Second, token probability displays inherent bias, tending to distribute within higher numerical ranges due to the decoding module's preference for selecting high-probability tokens. The CATP proposed in this study effectively addresses these deficiencies: First, its numerical distribution is no longer confined to high-value intervals; Second, for scenarios involving "incorrect responses with high token probability," this metric corrects the value to a lower magnitude, achieving alignment between confidence and correctness. On the other hand, a confidence boundary search module is designed to automate the partitioning of confidence distributions and articulate knowledge boundaries. Existing methods often rely on manual experience to set confidence thresholds for assessing LLM response reliability, resulting in high subjectivity and limited generalizability. The proposed confidence boundary search module automates interval partitioning of confidence distributions, employing a data-driven approach to determine boundaries corresponding to three knowledge mastery states: "Know", "Sciolism" and, "Unknow", thus circumventing limitations of manual threshold setting and enhancing the objectivity and generalization capability of knowledge boundary expression.

Compared to existing knowledge boundary expression research, this study introduces two core innovations and extensions: First, it incorporates a new dimension for identifying the "Sciolism" mastery state. This state specifically refers to an intermediate condition where the LLM lacks sufficient or complete knowledge required to answer a question, possessing only partial relevant knowledge. The addition of this dimension provides users with the option to reject LLM outputs under more conservative risk-averse scenarios involving partial knowledge. Second, the research context focuses on black-box LLMs. Most existing studies are conducted on white-box LLMs—where internal states are detectable and parameters accessible—with methodologies inherently reliant on these white-box characteristics. However, state-of-the-art LLMs are predominantly deployed as black-box services, characterized by non-disclosure of internal parameters, no fine-tuning support, and access restricted solely through API interfaces. The knowledge boundary expression problem for black-box LLMs remains underexplored and unresolved in current literature, despite its critical practical significance for deploying LLMs in reliable applications. This critical gap constitutes the primary motivation for our study, driving the development of knowledge boundary expression methods specifically designed for black-box LLMs.

## III. LSCL: A DEEP LEARNING-BASED METHOD FOR EXPRESSING KNOWLEDGE BOUNDARIES

For a given question $q_n$, the black-box LLM $M$ generates a response $\hat{A}$ and token probability $p_{\hat{A}}$ through internal computations, formally expressed as:

$$\hat{A}, p_{\hat{A}} = M(q_n)$$

The objective of this study is to develop a method that categorizes an LLM's knowledge state regarding question-answer capability into three distinct classes: "Know", "Sciolism" and, "Unknow".

This section presents LSCL, a deep learning-based method for expressing knowledge boundaries in black-box LLMs. The architecture of the proposed LSCL framework is illustrated in Figure 1. The methodological novelty resides in three key components: (1) Confidence estimation, (2) Confidence learning module, and (3) Adaptive thresholding module.

Within confidence estimation, inspired by the core principle of confidence calibration methods for black-box LLMs where "confidence aligns with correctness", this module computes CATP by integrating answer correctness with token probability. This calibrated confidence metric better reflects the model's knowledge mastery level and serves as the training target for the confidence learning module. The computational procedure is detailed in Section 3.1. The confidence learning module incorporates a question-answer alignment component that enhances local semantic relevance and global consistency between queries and LLM-generated responses, thereby improving confidence prediction accuracy. Its architectural design is elaborated in Section 3.2. Following confidence prediction, the adaptive thresholding module automatically determines upper and lower decision boundaries from training set confidence distributions to differentiate "Know", "Sciolism" and, "Unknow" states.

### A. Confidence Estimation: Correctness-Adjusted Token Probability

Token probability represent the output values generated by LLMs through modeling contextual information based on their parameterized knowledge, often regarded as the model's confidence in generating specific tokens (or answers). However, no reliable positive correlation exists between token probability and answer correctness: Higher token probability merely indicate greater confidence during generation, neither equating to mastery of required knowledge nor precluding incorrect answers; conversely, lower probabilities do not necessarily indicate complete knowledge absence, but may stem from insufficient confidence due to partial knowledge mastery. Therefore, the correctness of model responses serves as the essential criterion for determining genuine knowledge mastery of queried subjects.

Building upon this analysis, we propose CATP as a quantitative metric to more accurately assess models' mastery of knowledge required for correct responses. The computation of this metric involves two core procedures:

1. For a given input question $q$, the LLM generates an answer $\hat{A}$ using greedy decoding, obtaining the maximum token probability $p_{\hat{A}}$ at each generation step;

2. Incorporating correctness as a calibration factor through:

$$c = \begin{cases} p_{\hat{A}}, & \hat{A} = A \\ 1 - p_{\hat{A}}, & \hat{A} \neq A \end{cases} \tag{1}$$

where $A$ denotes the ground truth.

Compared to token probability, the CATP, by incorporating correctness information, can more accurately reflect the LLM's mastery of the knowledge required to answer questions. The differences and correspondences between the two can be intuitively observed in Figure 2(a). Specifically, by combining the level of token probability and the correctness of the answer, three types of knowledge mastery states for LLMs can be identified: 1) High CATP corresponds to high token probability values and correct answers, indicating that the LLM has mastered the correct knowledge required to answer the question; 2) Low CATP corresponds to high token probability values and incorrect answers, indicating that the LLM has acquired incorrect knowledge related to the question; 3) Medium CATP corresponds to low token probability values (regardless of whether the answer is correct or not), indicating that the LLM lacks sufficient knowledge to answer the question. The CATP is set as the prediction target for the confidence learning module.

It is noteworthy that since the CATP relies on the ground truth $A$ to compute correctness, this confidence score can only be utilized during the model training phase and cannot be directly applied during the inference phase (when the ground truth $A$ is unavailable). Furthermore, to categorize the model's knowledge mastery during inference into "Know", "Unknow", and the intermediate state of "Sciolism", it is necessary to automatically determine the partitioning thresholds $t_1$ and $t_2$ based on the confidence scores predicted during the training phase. This method will be detailed in Section 3.3.

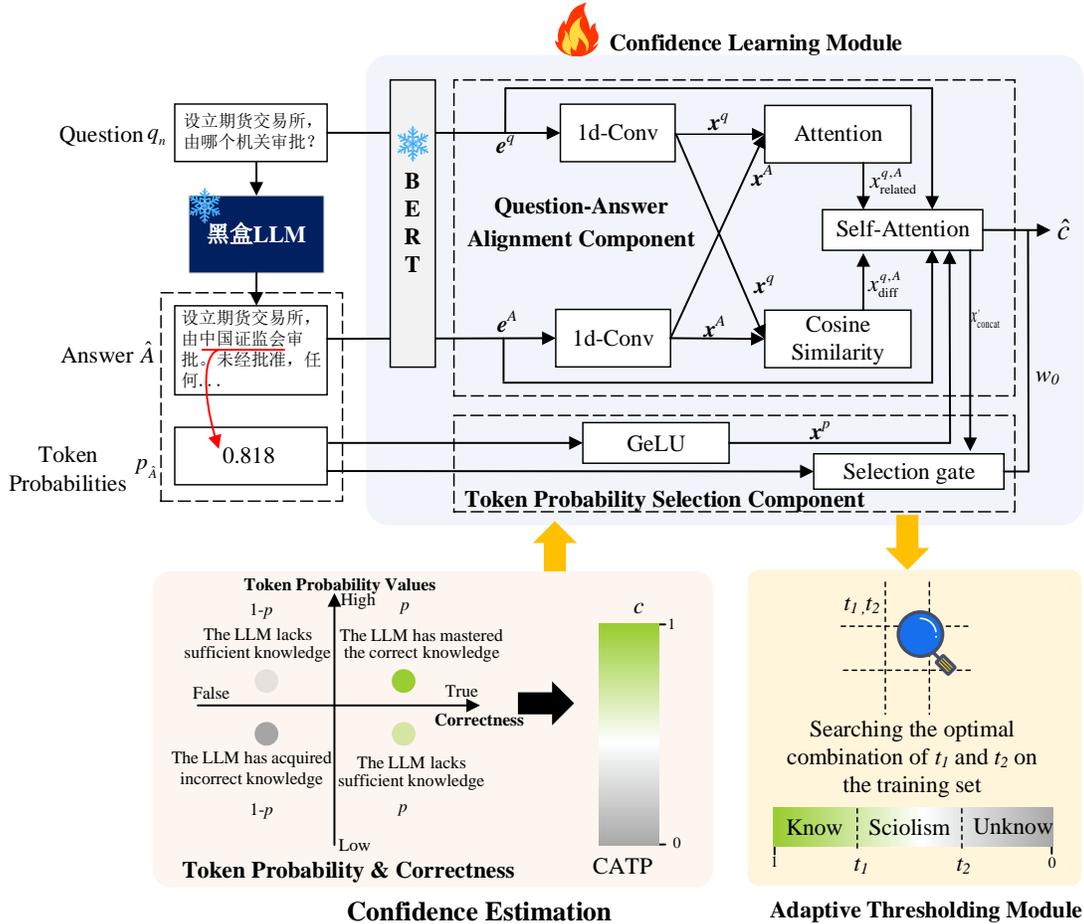

**Figure 1.** The procedure of LSCL

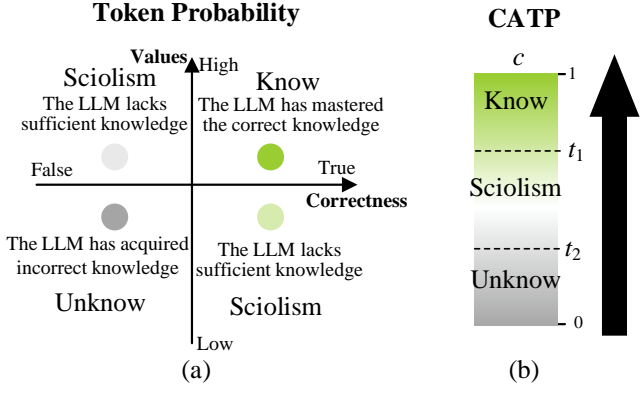

**Figure 2.** (a) Relationship between token probability-correctness quadrants and knowledge mastery states

### B. Confidence Learning Module

Upon establishing a confidence estimation methodology (CATP) for precise assessment of LLM knowledge mastery, we design a deep learning-based confidence learning module. This module achieves accurate target confidence prediction by learning intrinsic representations between inputs and outputs of black-box LLMs. The architecture embeds a knowledge distillation framework, where a parameter-efficient student model (confidence learning module) acquires internal knowledge representations from a large-scale teacher model (black-box LLM). In contrast to conventional knowledge distillation, two critical distinctions characterize our design: 1) The student model's core objective shifts from approximating the teacher's output probability distribution to predicting confidence scores reflective of knowledge states; 2) Requiring additional learning of deviation patterns between teacher outputs and ground-truth annotations, it directly quantifies knowledge mastery levels.

Consequently, the core functionality of the confidence learning module involves acquiring internal knowledge mappings from co-representations of black-box LLM inputs (questions) and outputs (answers, token probability) in multi-dimensional feature spaces, establishing correlations between these mappings and ground-truth confidence. Through constrained optimization, the module enforces the approximation of predicted confidence $\hat{c}$ to ground-truth confidence $c$ (values: $p_{\hat{A}}/1-p_{\hat{A}}$, see Section 3.1), ensuring distributional consistency between predicted and ground-truth confidence metrics.

The confidence learning module integrates two core components: a question-answer alignment component and a token probability selection component. Section 3.2 details the functionality and design rationale for each component. The question-answer alignment component employs BERT as its base encoder to transform questions and answers into distinct vector representations. Constructing a semantic matching mechanism aligns semantic relationships between question and answer vectors in the vector space, yielding a unified vector representation that captures the LLM's knowledge mastery and delivers core feature support for subsequent confidence prediction.

The token probability selection module generates gating weights based on the knowledge vector representation output by the question-answer alignment module, implementing dual constraints through specifically designed loss functions and penalty terms. Constraint one enforces the approximation of the predicted confidence $\hat{c}$ to ground-truth confidence $c$ (values: $p_{\hat{A}}/1-p_{\hat{A}}$), ensuring prediction accuracy. Constraint two imposes penalties on "ambiguous decision" behaviors—defined as predictions approaching the boundary threshold that hinder clear delineation of knowledge mastery states—enhancing discriminative power for classifying three types of knowledge mastery states.

**The Question-answer Alignment Component** The question-answer alignment component operationalizes the LLM's answer generation process, which essentially computes internal knowledge based on input questions. Latent semantic associations between generated answers and input questions map the LLM's knowledge mastery state regarding the target problem. This module processes question $q$ and the LLM generates an answer $\hat{A}$ as inputs, leveraging BERT to encode both into a shared feature space and generate corresponding vector representations: $e^q$ (question vector) and $e^A$ (answer vector). To comprehensively probe latent semantic relationships between $e^q$ and $e^A$ in the vector space, we subdivide semantic relation extraction into local and global dimensions, enabling multi-level semantic association capture.

For local semantic relation extraction, we apply a 1D convolutional layer to transform the question vector $e^q$ into $x^q$ and the answer vector $e^A$ into $x^A$, thereby extracting local features from both vectors:

$$x^q = *_{1d}\left(e^q, W_q, s, p\right) + b_q \tag{2}$$

$$x^A = *_{1d}\left(e^A, W_A, s, p\right) + b_A \tag{3}$$

Here, $W_q$ and $W_A$ denote weight matrices, $b_q$ and $b_A$ represent linear transformation biases, $s$ indicates the convolution kernel stride, $p$ specifies the input sequence padding length, and $*_{1d}(\cdot)$ corresponds to the 1D convolution operation.

We employ an attention mechanism to capture semantic interactions between $x^q$ and $x^A$:

$$x^{q,A}_{\text{related}} = \text{Attention}(x^q, x^A, \text{H}) \tag{4}$$

Here, $H$ denotes the number of attention heads and Attention($\cdot$) represents the multi-head attention mechanism. This mechanism employs multiple independent attention heads to capture localized dependencies within $x^q$ and $x^A$ through parallel computation, then integrates the results to model cross-space interactions.

Concurrently, we employ cosine similarity to quantify the discrepancy between $x^q$ and $x^A$, thereby measuring localized semantic divergence.

$$x^{q,A}_{\text{diff}} = \text{CosSim}(x^q, x^A) = \frac{(x^q, x^A)}{\|x^q\|_2 \cdot \|x^A\|_2} \tag{5}$$

Here, $\langle \cdot \rangle$ denotes the dot product between vectors, and $\|\cdot\|_2$ represents the Euclidean norm.

To extract global semantic relations, we concatenate $e^q$ and $e^A$ as input to the attention mechanism. For computational efficiency, we jointly feed the extracted local semantic relations along with global semantic vectors $e^q$ and $e^A$ into the attention mechanism to achieve feature alignment.

$$x_{concat} = concat(e^q, e^A, x_{related}^{q,A}, x_{diff}^{q,A}, x^p) \quad (6)$$

$$x'_{concat} = SelfAttention(x_{concat}, H) \quad (7)$$

Here, $concat(\cdot)$ denotes vector concatenation, and $SelfAttention(\cdot)$ represents the multi-head self-attention mechanism. $x^p$ denotes the token probability vector, which will be detailed in the next section.

**Token Probability Selection Component** As previously stated, the target confidence score is derived from the token probability of LLM-generated answers. This property introduces two fundamental challenges for confidence prediction modeling: 1. The confidence value carries physical interpretation, indicating the mastery level of knowledge required for LLM responses. Thus, the predicted confidence must closely approximate the ground-truth confidence (denoted as $p_{\hat{A}}$ or $1 - p_{\hat{A}}$) to accurately reflect the model's knowledge state. 2. When $p_{\hat{A}}$ approaches 0 or 1, the absolute error $\left| p_{\hat{A}} - \left(1 - p_{\hat{A}}\right) \right|$ amplifies significantly. Minor deviations in confidence may then cause substantial errors in knowledge state assessment, necessitating stricter constraints for predictions near the 0 and 1 boundaries.

To address these challenges, we design a token probability selection component. Its core mechanism employs a selection gate to generate a trust weight $w_0$ for the confidence learning module's reliance on token probability $p_{\hat{A}}$. This weight $w_0$ is then integrated into the loss function and penalty term design: a) The loss term forces predicted confidence toward $p_{\hat{A}}$ or $1 - p_{\hat{A}}$; b) The penalty term imposes stronger constraints on predictions near 0 or 1 to reduce boundary errors, thereby systematically resolving both challenges.

Within the token probability selection component, we initially embed the token probability value $p_{\hat{A}}$ of LLM responses into the deep learning model's vector space, transforming it into a vector representation $x_p$:

$$x^p = Gelu(W_p p_{\hat{A}}) + b_p \quad (8)$$

where $W_p$ denotes the weight matrix, $b_p$ the bias of the linear transformation, and $Gelu(\cdot)$ the activation function, with $x_p$ subsequently fed into the feature fusion layer, as specified in Equation (6).

Meanwhile, the token probability $p_{\hat{A}}$ undergoes a series of computations to derive the trust weight $w_0$ for the confidence learning module,

$$p'_{\hat{A}} = |2 \times p_{\hat{A}} - 1| \quad (9)$$

where $p_{\hat{A}}$ is transformed into $p'_{\hat{A}}$, exhibiting high input intensity when $p_{\hat{A}}$ approaches 0 or 1, and low otherwise. Subsequently, $p'_{\hat{A}}$ is converted into the trust weight $w_0$.

$$x''_{concat} = concat(p'_{\hat{A}}, x'_{concat}) \quad (10)$$

$$w_0 = Sigmoid(x''_{concat}) \quad (11)$$

where $Sigmoid(\cdot)$ represents the Sigmoid activation function, which maps $w_0$ to the interval $[0,1]$.

The question-answer alignment component module and token probability selection component compute and transform the questions, answers, and token probability into vector space representations within the model. Subsequently, the confidence prediction layer calculates the confidence prediction values:

$$\hat{c} = Sigmoid(x'_{concat}) \quad (12)$$

The confidence learning module generates the confidence prediction value $\hat{c}$ and trust weight $w_0$. We then design a loss function tailored to this specific problem:

$$loss = huber(c, \hat{c}, \delta) + \alpha \cdot mse(\hat{c}, \mu(p_{\hat{A}}, w_0)) + \beta \cdot \lambda(p_{\hat{A}}, w_0) \quad (13)$$

The function $huber(\cdot)$ is a loss function commonly used in regression problems. Adjusting the hyperparameter $\delta$ balances sensitivity to small errors and robustness to large deviations:

$$huber(c, \hat{c}, \delta) = f(x) = \begin{cases} \frac{1}{2}(c - \hat{c})^2, \ for |c - \hat{c}| \leq \delta \\ \delta \cdot \left(|c - \hat{c}| - \frac{1}{2}\delta\right), \ for |c - \hat{c}| > \delta \end{cases} \quad (14)$$

The second term of the loss function is designed to drive the output of the confidence learning module toward $p_{\hat{A}}$ or $1 - p_{\hat{A}}$. Here, $\mu(p_{\hat{A}}, w_0)$ quantifies the dynamic dependency between the trust weight $w_0$ and token probability $p_{\hat{A}}$, while $mse(\hat{c}, \mu(p_{\hat{A}}, w_0))$ denotes the loss function measuring the discrepancy between the confidence prediction and this dynamic dependency. Compared to directly computing the difference between $\hat{c}$ and $p_{\hat{A}}$, our method progressively guides the model predictions to approach $p_{\hat{A}}$ or $1 - p_{\hat{A}}$ without inducing optimization conflicts.

$$\mu(p_{\hat{A}}, w_0) = w_0 \cdot p_{\hat{A}} + (1 - w_0) \cdot (1 - p_{\hat{A}}) \quad (15)$$

The third loss term, $\lambda$, functions as a penalty designed to enforce conservative predictions for confidence score near 0 or 1. Specifically, $\lambda$ compels trust weight $w_0$ to converge toward 0 or 1 when $p_{\hat{A}}$ approaches these critical boundaries.

$$p'_{\hat{A}} = |2 \times p_{\hat{A}} - 1| \quad (16)$$

$$w'_0 = w_0 \cdot \log(w_0) + (1 - w_0) \cdot (1 - \log(w_0)) \quad (17)$$

$$\lambda(p_{\hat{A}}, w_0) = p_{\hat{A}} \cdot w'_0 \quad (18)$$

### C. Adaptive Thresholding Module

To express the knowledge boundaries of black-box LLMs, we discretize confidence predictions into three epistemic states: "Know", "Sciolism" and, "Unknow", thereby delineating knowledge boundaries. Let $\hat{y}_n$ denote the categorical label mapped from the confidence score $\hat{c}_n$, with this mapping formally defined as:

$$\hat{y}_n = \begin{cases} Know & , \hat{c}_n \geq t_1 \\ Sciolism & , t_2 < \hat{c}_n < t_1 \\ Unknow & , \hat{c}_n \leq t_2 \end{cases} \quad (19)$$

where $t_1$ and $t_2$ represent the upper and lower demarcation thresholds, respectively.

To identify optimal upper/lower demarcation thresholds $(t_1, t_2)$ for confidence predictions on the training set $\hat{C}_{train}$, we develop a grid search-based boundary optimization algorithm. This frames threshold optimization as a search problem where the algorithm seeks $(t_1, t_2)$ that maximizes the Macro-$F_1$ score($F_1$), which quantifies the classification performance of labels partitioned by each threshold combination on $\hat{C}_{train}$.

The detailed computational procedure is implemented as follows:

**Algorithm 1** Grid Search-based Boundary Optimization Algorithm

**Input:** Let $\hat{C}_{train} = \{(\hat{c}_n, y_n) | n = 1, 2, \ldots, N_{train}\}$ denote the training set, where $\hat{c}_n \in [0,1]$ represents the prediction confidence of the $n$-

th sample, $y_n \in \{$"Know", "Sciolism", "Unknow"$\}$ is the manually annotated label, and $N_{\text{train}}$ is the total number of samples in $\hat{C}_{\text{train}}$. The parameter $m_{\text{bins}}$ indicates the number of uniformly partitioned candidate boundaries.

**Output:** The upper and lower demarcation thresholds $(t_1, t_2)$:

1:  $F_1^* = -1$; // Initialize the objective function value.
    // Generate candidate sets for upper bound $t_1$ and lower
    bound $t_2$ through uniform partitioning.
2:  $L = 0, U = 1, \Delta = (U - L)/(m_{\text{bins}} - 1)$;
3:  $\mathrm{T}_1 = \{U - k \cdot \Delta_1 | k = 1, \dots, m_{\text{bins}} - 1\}$,
    $\mathrm{T}_2 = \{L + k \cdot \Delta_2 | k = 1, \dots, m_{\text{bins}} - 1\}$;
    // Grid search
4:  **for** $t_1^* \in \mathrm{T}_1, t_2^* \in \mathrm{T}_2$ **do**
5:      **if** $t_2^* \leq t_1^*$ **then**
            // Classify training samples
6:          $Y_{\text{know}}(t_1^*) = \{BE(\hat{c}_n) | \hat{c}_n \geq t_1^*, \hat{c}_n \in \hat{C}_{\text{train}}\}, Y_{\text{unkonw}}(t_2^*) = \{BE(\hat{c}_n) | \hat{c}_n \leq t_2^*, \hat{c}_n \in \hat{C}_{\text{train}}\}$;
7:          $Y_{\text{sciolism}}(t_1^*, t_2^*) = \{BE(\hat{c}_n) | t_2^* < \hat{c}_n < t_1^*, \hat{c}_n \in \hat{C}_{\text{train}}\}$;
8:          **Compute** the $F_1^*$;
9:          **if** $F_1^* \geq F_1$ **then**
10:             $F_1 = F_1^*, t_1 = t_1^*, t_2 = t_2^*$;
11:         **end if**
12:     **end if**
13: **end for**
14: **return** $t_1, t_2$

## IV. Experimental Setup

### A. Datasets and LLMs

To validate the cross-domain efficacy of our proposed method, two domain-specific datasets were employed:

- MMedBench [23]: A comprehensive multilingual medical competency benchmark curated from national medical licensing examinations. Only the Chinese-language subset was utilized, comprising 26,000 training and 3,000 validation samples;
- CFLUE [4]: A Chinese Financial Language Understanding Evaluation benchmark for assessing LLM capabilities in financial contexts. We selected its multiple-choice and true/false subsets to facilitate token probability extraction (30,000 training and 10,000 validation instances).

To validate the efficacy and generalizability of the proposed method, three representative LLMs were treated as black-box systems: including one open-source model (Qwen2-7B) and two proprietary systems (Qwen-flash, Doubao-flash). The preprocessing pipeline comprised three stages: First, each query was fed to all three LLMs to capture the answer $\hat{A}$ with corresponding token probability $p_{\hat{A}}$; Subsequently, $p_{\hat{A}}$ were transformed into CATP $c$ using the confidence metric defined in Section 3.1; Finally, knowledge mastery labels were assigned via an interval partitioning strategy.

The labeling protocol operated as follows: Equal-width binning partitioned confidence scores $c$ into 10 intervals over $[0,1]$, with accuracy rates computed per bin. Instances in bins exceeding 90% accuracy received "Know" labels, those below 20% were labeled "Unknow", and intermediate bins were assigned "Sciolism". This consistent labeling schema across datasets ensured experimental fairness and comparability. Notably, threshold values (90%/20%) remain adjustable: For high-stakes domains (e.g., medical/financial consulting), the threshold of "Know" may be elevated to mitigate operational risks.

### B. Baselines

To demonstrate the superiority of our approach, two prevalent baseline methodologies compatible with black-box LLMs were evaluated: confidence-based approaches and prompting-based techniques. The conceptual frameworks and implementations are detailed below.

**Based Confidence Methods** The core of this approach is to obtain quantifiable confidence scores (e.g., token probability) from the model output. An optimal threshold is then determined on a labeled training dataset, which converts the continuous scores into a multi-class determination of "knows," "does not know," and "partial knowledge." Specific implementations include two schemes:

- **Guiding by Prompt**: Following the method proposed by Tian et al.[20], we guide the LLM to output both the answer to the question and its confidence score by designing specific prompts. The prompt used is: "Q: {question} Please answer this question and provide your confidence level."
- **Token probability**: By inputting a question with constrained options to the LLM, the model's output range is restricted to accurately obtain the token probability at the output layer. The prompt employed is: "Q: {question} Select the most appropriate answer from the following options: A.{option A}, B.{option B}, C.{option C}, D.{option D}." The feasibility of this scheme arises from the fact that the majority of mainstream black-box LLMs support access to token probability at the output layer.

**Based Prompt Method** The core of this method involves using prompts to enable the model to articulate its knowledge boundaries in natural language. Specifically, it encompasses two implementation schemes:

- **Prior Prompt:** Drawing on the research of Ren et al.[7], this approach assesses the model's knowledge state regarding a question by prompting it upfront to indicate whether it is willing to decline to answer. The prompt employed is: "Do you honestly know the answer to the following question? If you know, output 'Yes'; otherwise, output 'No'. Respond with only a single word: 'Yes' or 'No'. Q: {question}";
- **Posterior Prompt**: Building on the research of Ren et al.[7] and Kadavath et al.[24], this method queries the model's confidence in its answer after the LLM has generated a response. The prompt used is: "Are you confident that the answer to the following question is correct? Q: {question} A: {answer} If confident, output 'Confident'; otherwise, output 'Unconfident'. Respond with only a single word: 'Confident' or 'Unconfident'."

### C. Implementation Details

The confidence learning module was trained on a consumer-grade GPU equipped with an NVIDIA RTX 4060-Ti-16GB. Key hyperparameter settings are detailed in Table 1. Observation revealed a highly imbalanced frequency distribution in the true confidence scores output by all LLMs: the vast majority of samples exhibited confidence levels approaching 1, while the number of samples within low-confidence intervals was extremely scarce. This distribution characteristic indicates that data imbalance constitutes a primary challenge for achieving high-accuracy confidence prediction. If unaddressed, the model's predictive performance on low-confidence samples may degrade significantly, thereby compromising the reliability and generalization capability of the overall prediction results.

To mitigate the data imbalance issue, we implemented the following sample augmentation strategy: First, the confidence range [0,1] is uniformly partitioned into 10 intervals. Next, the sample count within each interval is computed, and intervals with counts below the median across all intervals are designated for augmentation. Finally, samples within these designated intervals are oversampled using the SMOTE method to generate new synthetic samples.

**Table 1.** Configuration of Primary Hyperparameters

| Module | Hyperparameters | Values |
|---|---|---|
| Confidence Learning Module | Initial learning rate | 0.0005 |
| | Dropout rate | 0.1 |
| | Weight decay rate | 0.0001 |
| | Optimizer | AdamW |
| | $\delta$ | 0.15 |
| | $\alpha$ | 0.5 |
| | $\beta$ | 0.1 |
| Adaptive Thresholding Module | $m_{bins}$ | 100 |

### D. Training Performance

This section demonstrates the performance of the confidence learning module on the training set and the search for the optimal confidence boundary using the MMedBench dataset as an example. Analysis of CFLUE is provided in Appendix 1. The comparative analysis of the confidence learning module's predictive performance on the training set (see Figures 3-5) provides key evidence for the model's effectiveness. The frequency distributions of predicted confidence versus true confidence across the three LLMs reveal a strong agreement between the predicted and true distributions. This indicates that the model effectively captures the underlying relationships between question information, answer information, token probability, and predicted confidence, demonstrating reliable predictive capability. Further examination of the residual distribution plots (predicted minus true confidence) shows that the histograms for all three LLMs indicate that the residuals for the vast majority of samples are clustered near zero. This finding further validates the superior performance of the confidence learning module, indicating low prediction bias and high accuracy in confidence estimation.

Following training, the optimal upper boundary $(t_1)$ and lower boundary $(t_2)$ were determined by the adaptive thresholding module. The specific results are visualized in Figure 6. Experimental results demonstrate that effective and robust optimal boundary parameters were obtained for all three LLMs on the MMedBench dataset. These parameters successfully partitioned the samples into three distinct categories: "Know" $(\hat{c} \geq t_1)$, "Sciolism" $(t_1 < \hat{c} < t_2)$, and "Unknow" $(\hat{c} \leq t_2)$.

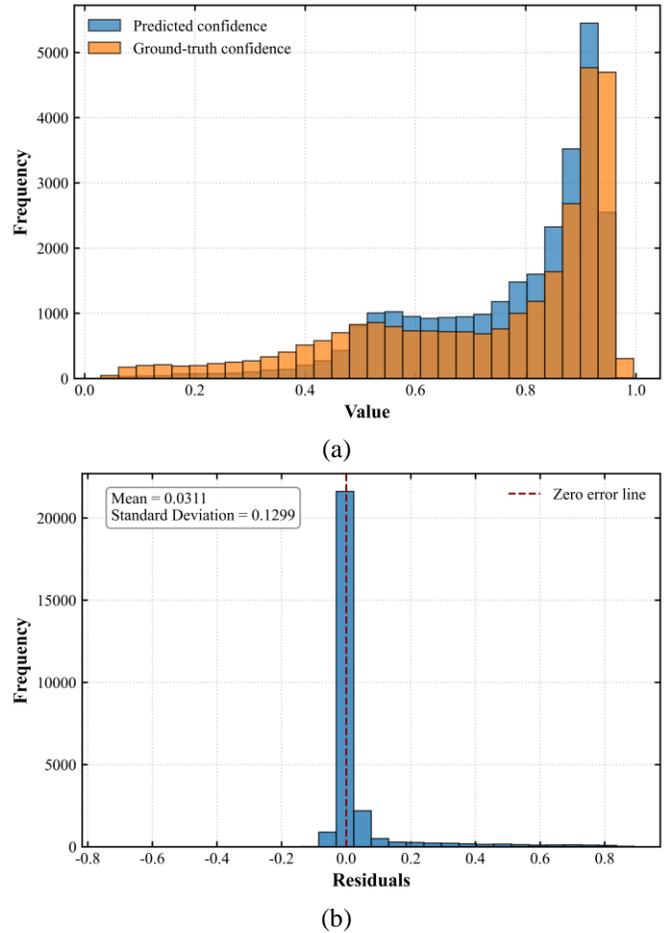

(a)

(b)

**Figure 3.** The confidence learning module for Qwen2-7B's: (a) frequency distribution of predicted confidence versus true confidence; (b) residual distribution histogram.

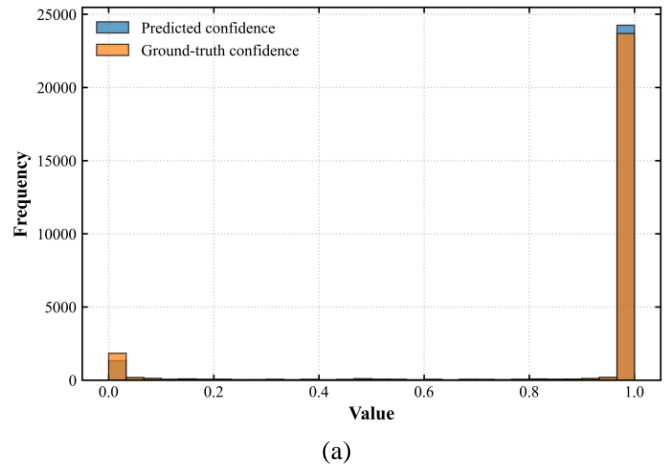

(a)

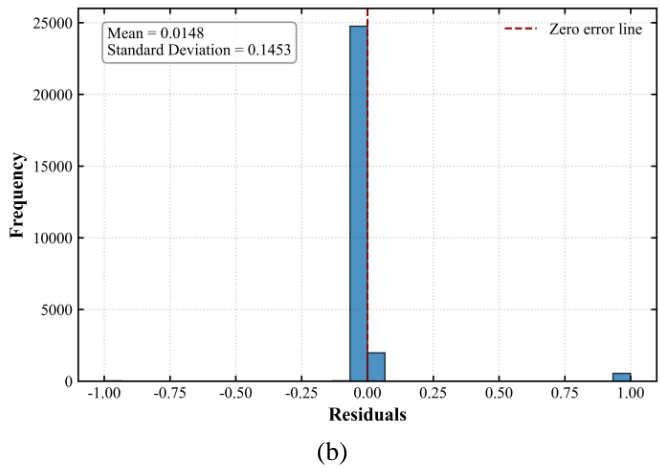

**Figure 4.** The confidence learning module for Qwen-flash's: (a) frequency distribution of predicted confidence versus true confidence; (b) residual distribution histogram.

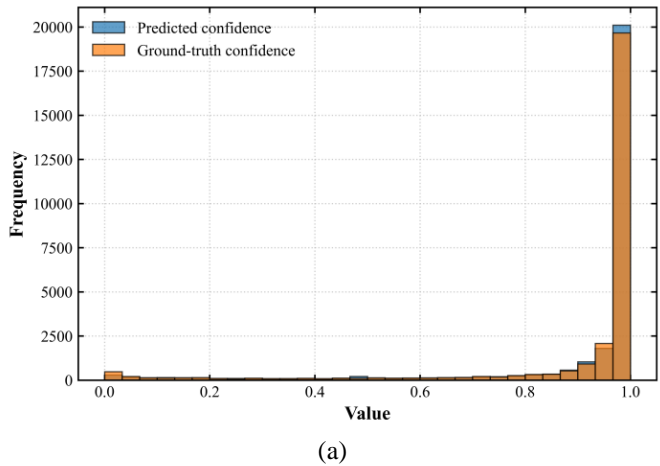

(a)

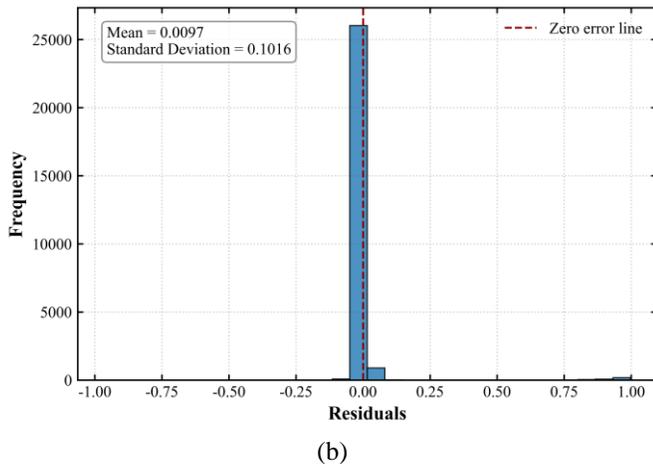

(b)

**Figure 5.** The confidence learning module for Doubao-flash's: (a) frequency distribution of predicted confidence versus true confidence; (b) residual distribution histogram.

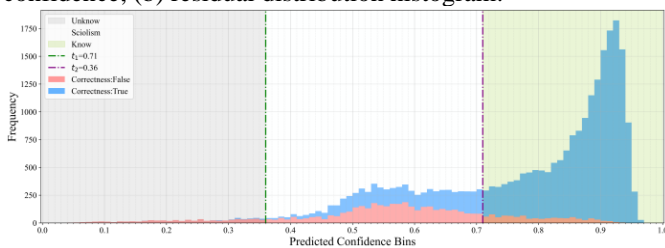

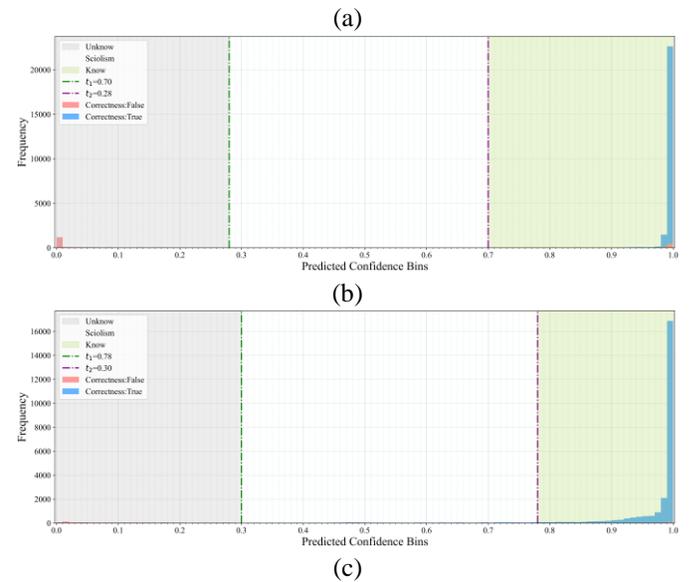

**Figure 6.** Histograms of predicted confidence by LSCL on the MMedBench training set: (a) Qwen2-7B, (b) Qwen-flash, and (c) Doubao-flash.

### E. Overall Performance

After determining the optimal confidence boundaries on the training set, Table 2 presents the main experimental results on the MMedBench and CFLUE datasets. The key findings are as follows:

Across all tested LLMs, the proposed LSCL method significantly outperformed all baseline models, demonstrating particularly strong performance on the $F_1$ and recall rate metrics. Specifically:

- On Qwen2-7B: LSCL achieved an $F_1$ of 0.696 and a recall rate of 0.696, whereas the best-performing baseline model attained only 0.650 in $F_1$ and 0.665 in the recall rate.

- On Qwen-flash: LSCL achieved an $F_1$ of 0.670 and a recall rate of 0.675, compared to the best baseline model's maximum of 0.619 in $F_1$ and 0.610 in the recall rate.

- On Doubao-flash: LSCL achieved an $F_1$ of 0.699 and a recall rate of 0.691, while the best baseline model achieved at most 0.648 in $F_1$ and 0.668 in the recall rate.

Regarding accuracy, while LSCL did not achieve the top performance across all LLMs, it consistently outperformed the majority of baseline methods. Notably, the accuracy of all models exceeded their corresponding $F_1$ and recall rates. This phenomenon arises from a general deficiency in existing methods to effectively classify samples representing "Unknow" and "Sciolism" categories, which aligns with the observations reported by Chen et al. [8]. Crucially, the proposed LSCL method achieves state-of-the-art performance on this specific classification task among existing approaches. This observation can be explained from a deep learning theoretical perspective: the number of samples belonging to the "Know" category significantly outweighs those in the "Unknow" and "Sciolism" categories. This imbalance induces a majority class bias during model training, fundamentally manifesting as a data imbalance problem. Given that the confidence learning module within LSCL is inherently a deep learning model, developing targeted strategies to mitigate data imbalance on this foundation holds promise for further enhancing the model's discriminative ability across all sample categories, thereby enabling more precise delineation of knowledge boundaries.

**Table 2.** Performance comparison between LSCL and baseline models on the MMedBench and CFLUE test sets.

| | Method | MMedBench | | | CFLUE | | |
|---|---|---|---|---|---|---|---|
| | | $F_1$ | Accuracy | Recall | $F_1$ | Accuracy | Recall |
| Qwen2-7B | Guiding by prompt | 0.262 | 0.648 | 0.333 | 0.367 | 0.504 | 0.389 |
| | Token probs | 0.650 | **0.940** | 0.665 | 0.525 | 0.768 | 0.539 |
| | Prior prompt | 0.462 | 0.772 | 0.492 | 0.400 | 0.651 | 0.500 |
| | Posterior prompt | 0.459 | 0.794 | 0.500 | 0.444 | 0.447 | 0.484 |
| | **LSCL** | **0.696** | 0.919 | **0.691** | **0.687** | **0.862** | **0.678** |
| Qwen-flash | Guiding by prompt | 0.312 | 0.881 | 0.333 | 0.061 | 0.099 | 0.334 |
| | Token probs | 0.619 | 0.903 | 0.610 | 0.314 | 0.467 | 0.503 |
| | Prior prompt | 0.488 | 0.878 | 0.502 | 0.462 | 0.744 | 0.487 |
| | Posterior prompt | 0.096 | 0.106 | 0.500 | 0.449 | 0.801 | 0.502 |
| | **LSCL** | **0.670** | **0.907** | **0.675** | **0.678** | **0.769** | **0.706** |
| Doubao-flash | Guiding by prompt | 0.395 | 0.863 | 0.400 | 0.279 | 0.437 | 0.364 |
| | Token probs | 0.648 | **0.944** | 0.668 | 0.572 | **0.843** | 0.650 |
| | Prior prompt | 0.526 | 0.917 | 0.523 | 0.506 | 0.785 | 0.515 |
| | Posterior prompt | 0.539 | 0.919 | 0.531 | 0.466 | 0.756 | 0.515 |
| | **LSCL** | **0.699** | 0.940 | **0.691** | **0.644** | 0.819 | **0.668** |

**Table 3.** Comparison of LLM answer accuracy rates between LSCL prediction and observed values.

| **MMedBench** | | | | |
|---|---|---|---|---|
| | | Know(%) | Sciolism (%) | Unknow(%) |
| Qwen2-7B | Truth labels | 99.59(2220) | 53.91(1011) | 2.05(195) |
| | Prediction labels ($t_1 = 0.71, t_2 = 0.36$) | 93.61(2394) | 54.75(1007) | 58.73(25) |
| Qwen-flash | Truth labels | 100.00(3026) | 45.78(317) | 0.00(83) |
| | Prediction labels ($t_1 = 0.70, t_2 = 0.28$) | 97.74(3260) | 45.26(96) | 40.85(70) |
| Doubao-flash | Truth labels | 99.90(2986) | 64.80(250) | 3.16(190) |
| | Prediction labels ($t_1 = 0.78, t_2 = 0.30$) | 95.70(3112) | 45.26(250) | 40.85(64) |
| **CFLUE** | | | | |
| | | Know(%) | Sciolism (%) | Unknow(%) |
| Qwen2-7B | Truth labels | 99.99(941) | 54.58(1489) | 6.56(264) |
| | Prediction labels ($t_1 = 0.76, t_2 = 0.28$) | 99.72(1028) | 48.38(1571) | 68.42(100) |
| Qwen-flash | Truth labels | 99.95(1952) | 69.92(476) | 0.00(266) |
| | Prediction labels ($t_1 = 0.71, t_2 = 0.30$) | 83.79(2005) | 68.64(338) | 64.10(351) |
| Doubao-flash | Truth labels | 99.80(2019) | 50.54(370) | 3.28(305) |
| | Prediction labels ($t_1 = 0.79, t_2 = 0.30$) | 91.37(2028) | 49.08(434) | 62.93(232) |

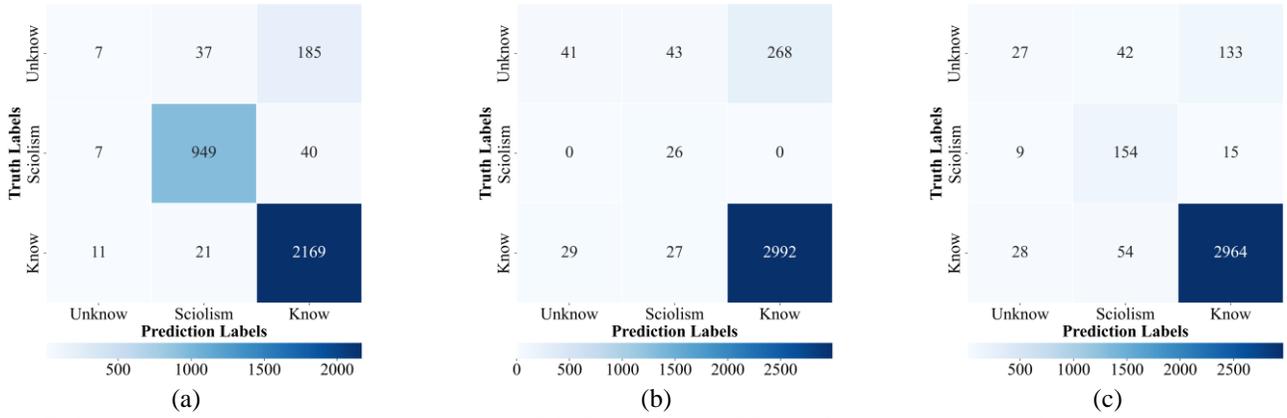

**Figure 7.** Confusion matrices for prediction labels by LSCL: (a) Qwen2-7B, (b) Qwen-flash, and (c) Doubao-flash

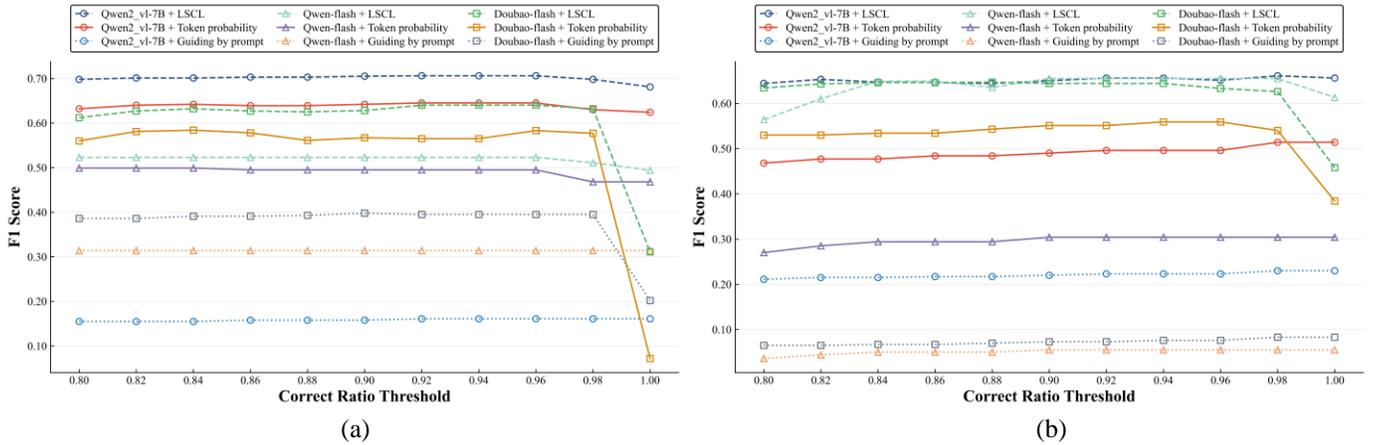

**Figure 8.** Comparison of knowledge boundary expression performance under different accuracy thresholds on datasets (1) MMedBench and (2) CFLUE.

For further analysis, we conducted a numerical analysis of the LSCL knowledge boundary labels using the MMEdBench dataset as an example, with the results presented in Figure 7. The results demonstrate that LSCL effectively replaces the original LLM to achieve precise delineation of knowledge boundaries, attaining an accuracy exceeding 0.907. This superior performance is primarily attributed to the efficient and accurate classification of samples in the "Know" and "Sciolism" categories. However, the method exhibits significant limitations in classifying samples of the "Unknow" category, where some "Unknow" samples are misclassified as "Know". A deeper investigation reveals that token probability inputs enable the confidence learning module to predict numerical confidence ranges effectively; however, the learning process for low-confidence samples necessitates sufficient sample size support. For instance, given a sample with a token probability of 0.3, its CATP can only be 0.3 or 0.7. A data imbalance problem biases the model towards outputting high-confidence results, and 0.3 and 0.7 are likely assigned to distinct classification labels. Although sample sizes for the "Sciolism" category are similarly limited, data imbalance exerts a comparatively minor impact on predictions within this range. For example, a sample with a token probability of 0.6 may yield confidence values of 0.4 or 0.6, both potentially falling within the "Sciolism" interval.

Furthermore, the relationship between LSCL knowledge boundary labels and answer correctness is presented in Table 3. The minimum accuracy of "Know" category samples classified by the confidence boundary search module is 93.61%, which exceeds the 90% accuracy threshold set during the dataset partitioning stage. This empirical result validates the effectiveness and reliability of the proposed method in predicting "Know" category samples. In contrast, the maximum difference between the predicted accuracy of "Unknow" category samples and the preset 30% accuracy requirement reaches 28.73%, indicating a significant performance gap. It should be noted that the suboptimal prediction performance for "Unknow" samples is not a specific issue of the proposed LSCL method, but a common challenge faced by all current knowledge boundary identification methods [25].

### F. Performance of Models with Different Accuracy Rates

In high-risk scenarios such as healthcare and finance, where seeking LLM professional advice requires caution, the "Know" status may demand higher accuracy thresholds to mitigate decision risks. However, existing studies lack discussion on the performance of knowledge boundary expression methods under varying accuracy thresholds. To address this, we established accuracy thresholds ranging from 0.8 to 1.0 in increments of 0.02, evaluating the performance of diverse knowledge boundary expression methods across these thresholds, with results shown in Figure 8. As accuracy thresholds increase, all models exhibit varying degrees of performance fluctuation. Notably, when the accuracy threshold is set to 1.0, a significant performance degradation occurs. Nevertheless, the proposed LSCL outperforms baseline models across all threshold configurations.

### G. Performance Evaluation of Adaptive Alternative Solutions When Token Probability is Unavailable

Certain black-box LLMs accessible via API, such as DeepSeek-3.2 and ChatGPT-4, lack token probability output capability, rendering the token probability-dependent LSCL method inapplicable. More specifically, token probability cannot be utilized for confidence computation. To address this limitation, we propose a compatibility-focused alternative: Prompt-guided LSCL. This approach substitutes token probability with confidence scores derived from model-generated tokens, implemented through the following procedure: 1. Design prompts to elicit both answers $\hat{A}$ and the token confidence scores $s_{\hat{A}}$ from LLMs. The implemented prompt template is: "Q: {question} Please answer and provide your confidence level"; 2. Apply the logic of Equation (1) to calibrate the token confidence scores $s_{\hat{A}}$, obtaining calibrated confidence metrics $c$.

The Prompt-guided LSCL method was evaluated on two public datasets and three large language models (LLMs), with results detailed in Table 4. Compared to the prompt-only guided baseline, Prompt-guided LSCL demonstrated a significant performance improvement of at least 37%. Despite exhibiting marginally inferior performance relative to the original LSCL in most scenarios, the performance gap was minimal, and Prompt-guided LSCL achieved superior results in specific cases. These findings indicate that Prompt-guided LSCL serves as a reliable adaptive alternative for knowledge boundary expression in LLMs where token probability are inaccessible.

Furthermore, the guiding by prompt exhibited substantial instability in evaluating Qwen-Flash's knowledge boundary expression on the CFLUE dataset, achieving $F_1$ of merely 0.061. Nevertheless, after processing through our proposed methodology, Prompt-guided LSCL demonstrated robust performance with $F_1$ of 0.678. This contrast further validates the robustness and effectiveness of the proposed approach.

**Table 4.** Performance comparison between **Prompt-guided LSCL** and baseline models on MMedBench and CFLUE test sets

| | Method | MMedBench | CFLUE |
|---|---|---|---|
| Qwen2-7B | Guiding by prompt | 0.262 | 0.367 |
| | **Prompt-guided LSCL** | 0.595 | **0.703** |
| | **LSCL** | **0.696** | 0.687 |
| Qwen-flash | Guiding by prompt | 0.312 | 0.061 |
| | **Prompt-guided LSCL** | 0.531 | 0.624 |
| | **LSCL** | **0.670** | **0.678** |
| Doubao-flash | Guiding by prompt | 0.395 | 0.279 |
| | **Prompt-guided LSCL** | 0.633 | 0.593 |
| | **LSCL** | **0.699** | **0.644** |

## V. Conclusion

This study focuses on black-box LLMs with extensive parametric knowledge but inaccessible internal parameters, highlighting the critical research value of establishing explicit knowledge boundaries for such models. To address this, we propose LSCL—a deep learning-based knowledge boundary expression method—designed to assess an LLM's mastery of knowledge required for generating responses. Using correctness-adjusted token probability as ground-truth confidence scores, we developed a deep neural network architecture inspired by knowledge distillation. This model learns to map LLM inputs (questions) and outputs (answers and token probability) to the model's internal knowledge mastery state regarding queried topics. Furthermore, LSCL incorporates a confidence boundary search module that categorizes knowledge mastery levels into three states: "Know", "Sciolism" and, "Unknow".

Experimental results demonstrate that LSCL achieves high-precision confidence prediction and effectively delineates knowledge boundaries in LLMs. This study underscores the necessity of knowledge boundary expression for black-box LLMs and delivers a practical framework for its implementation. For black-box LLMs lacking token probability access, we introduce an adaptive alternative that approaches LSCL's performance while significantly outperforming baseline models. Furthermore, addressing stringent reliability requirements in high-stakes domains (e.g., healthcare and finance), we validate the method's robustness across varying accuracy thresholds. LSCL consistently outperforms baselines in all scenarios.

## VI. Appendix 1: On CFLUE Dataset Training Performance of Confidence Learning Module and Optimal Confidence Boundary Search Results

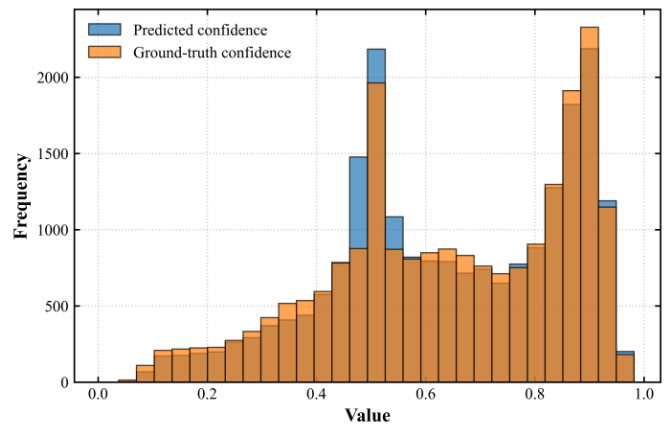

(a)

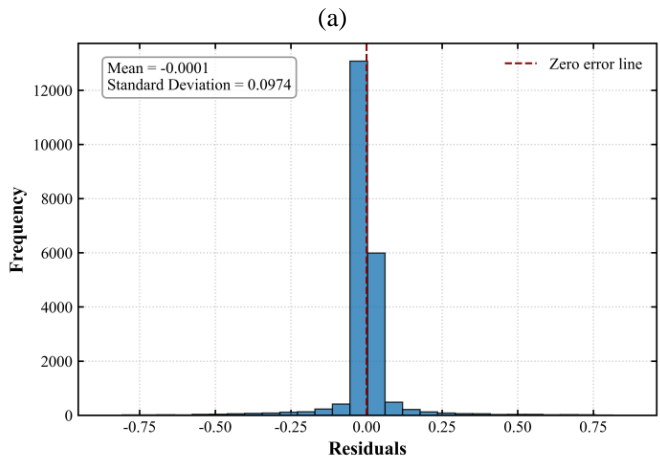

(b)

Figure 9. The confidence learning module for Qwen2-7B's: (a) frequency distribution of predicted confidence versus true confidence; (b) residual distribution histogram.

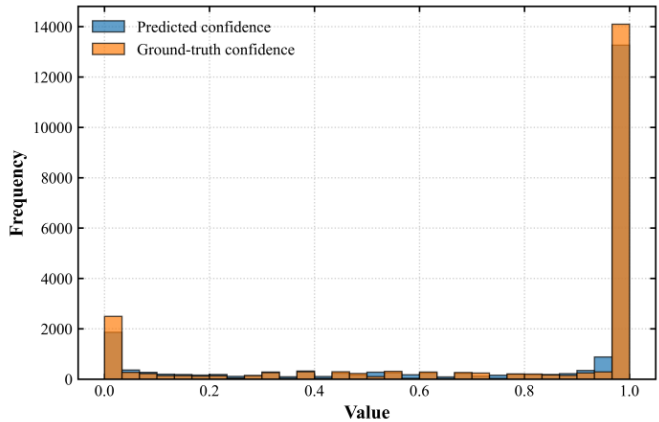

(a)

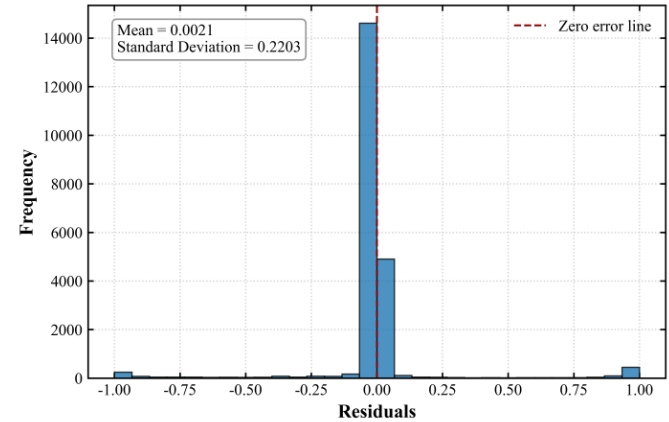

(b)

Figure 10. The confidence learning module for Qwen-flash's: (a) frequency distribution of predicted confidence versus true confidence; (b) residual distribution histogram.

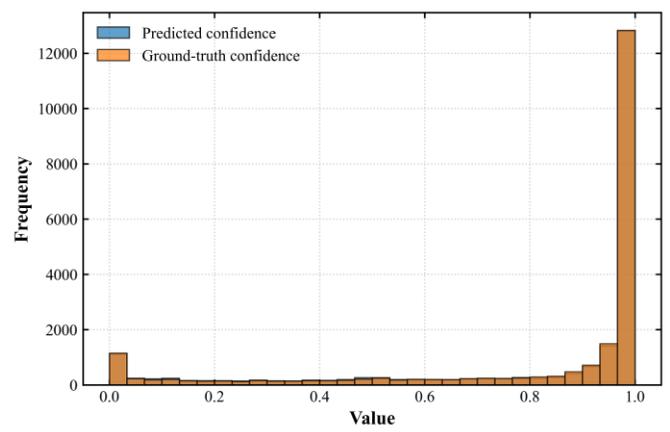

(a)

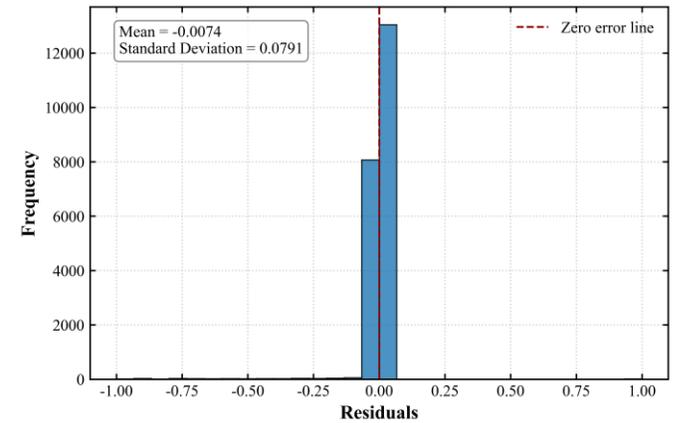

(a)                                                    (b)

Figure 11. The confidence learning module for Doubao-flash's: (a) frequency distribution of predicted confidence versus true confidence; (b) residual distribution histogram.

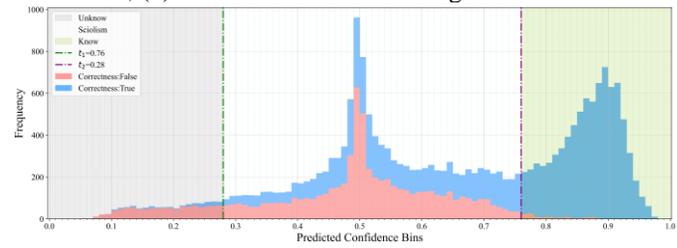

(a)

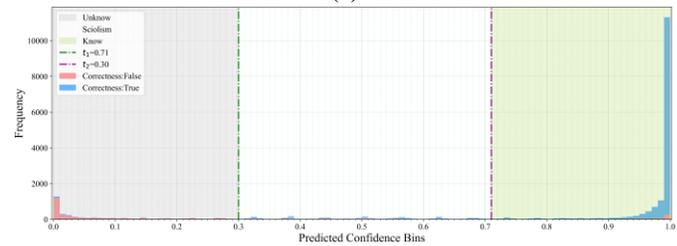

(b)

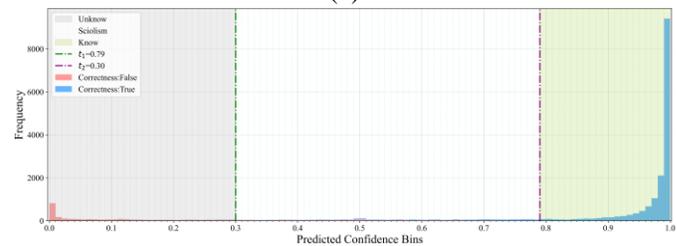

(c)

Figure12. Histograms of predicted confidence by LSCL on the CFLUE training set: (a) Qwen2-7B, (b) Qwen-flash, and (c) Doubao-flash.

**Haotian Sheng** was born in Ningxia province, China in 1997. He received a Master's degree in Logistics engineering and management from Dalian Maritime University, Dalian, Liaoning, China in 2022. He is currently pursuing a Doctor's degree in Management Science and Engineering at South China University of Technology, Guangzhou, Guangdong, China. His research interests include the application of LLM in advertising regulation based on the method of RAG, fine-tuning, and prompt engineering.

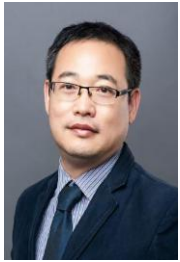

**Heyong Wang**, professor and doctoral supervisor of South China University of technology, whose research fields cover data mining, big data, business intelligence, etc. He has visited the University of Miami in the United States, presided over a number of scientific research projects, published many papers, won many honors, and actively participated in social part-time work.

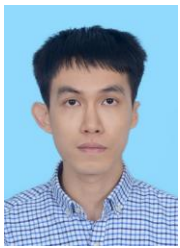

**Ming Hong**, assistant professor, master supervisor and doctor of management, Department of e-commerce, South China University of technology. The main research directions are data mining, text mining, business intelligence and recommendation system, and many papers have been published. It has long focused on the application and innovation of artificial intelligence technology in the field of digital commerce, and is committed to promoting the research of business intelligence and decision support system driven by data.

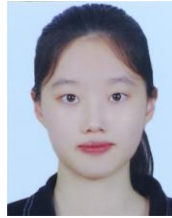

**Hongman He**, from Shenzhen, Guangdong, China, is currently pursuing a PhD in Management Science and Engineering at South China University of Technology in Guangzhou, Guangdong Province, China. She focuses on multimodal deep forgery detection and defense research, dedicated to providing practical technical solutions for platform content security governance, including but not limited to social media, video platforms, news media, etc., helping platform managers quickly identify and process forged content, and maintaining a healthy network ecosystem.

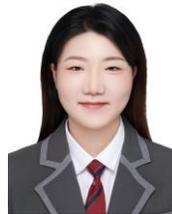

**Junqiu Liu** received the Bachelor of Management degree from the Department of Electronic Commerce, South China University of Technology, Guangzhou, China, in 2023. She is currently working toward the Doctor of Philosophy degree in Management Science and Engineering with the Department of Electronic Commerce, South China University of Technology, Guangzhou, China. Her research interests include fact-checking based on Retrieval-Augmented Generation (RAG) and Large Language Model multi-agents in the short video scenario.